\documentclass{article}

\usepackage{spconf,amsmath,epsfig}

\usepackage{spconf,amsmath,graphicx}
\usepackage{subfig}
\usepackage{color,soul}
\usepackage{epstopdf}
\usepackage{tikz}
\usepackage{balance}
\usepackage{url}
\usepackage{multirow,hhline}
\usepackage{placeins}
\usepackage{wasysym}
\usepackage{amssymb}

\usepackage[nomessages]{fp}% http://ctan.org/pkg/fp
\newcommand{\bx}{\mathbf{x}}

\newcommand{\bz}{\mathbf{z}}

\newcommand{\hx}{\hat{\mathbf{x}}}
\newcommand{\hy}{\hat{\mathbf{y}}}

\newcommand{\tx}{\widetilde{\mathbf{x}}}
\newcommand{\xH}{\widetilde{\mathbf{x}}_{\rm H}}
\newcommand{\zH}{\mathbf{z}_{\rm H}}

\newcommand{\image}{\pgfuseimage}

\newcommand{\figpath}{./Figures/}
\newcommand{\pngBIG}{./PNG/Result_Paper_Entire/}
\newcommand{\pngDET}{./PNG/Result_Paper_Mazzini/}
\newcommand{\pngFULL}{./PNG/Result_Paper_Mazzini_Full/}
\definecolor{colPeppe}{rgb}{0.7, 0.2, 0.15}

\definecolor{colSico}{rgb}{0.9, 0.5, 0.1}

\pgfdeclareimage[width=0.9\textwidth]{net2}{\figpath flowchart.pdf}

\newcommand{\widBIG}{3.2cm}
\newcommand{\widDET}{2.4cm}

\pgfdeclareimage[width=\widBIG]{Bx}{\pngBIG 004_RGBDR_RGB.png}
\pgfdeclareimage[width=\widBIG]{Bz}{\pngBIG 004_RGBD_RGB.png}
\pgfdeclareimage[width=\widBIG]{Br}{\pngBIG 004_RGB_RGB.png}
\pgfdeclareimage[width=\widBIG]{Bhpf}{\pngBIG 004_RGB__HPF_RGB.png}
\pgfdeclareimage[width=\widBIG]{Bfive}{\pngBIG 004_RGB_PANIGARSS_RGB.png}
\pgfdeclareimage[width=\widBIG]{Bprop}{\pngBIG 004_RGB_PANS5HBTY_RGB.png}

\pgfdeclareimage[width=\widDET]{Dx}{\pngDET 004_RGBDR_RGB.png}
\pgfdeclareimage[width=\widDET]{Dz}{\pngDET 004_RGBD_RGB.png}
\pgfdeclareimage[width=\widDET]{Dr}{\pngDET 004_RGB_RGB.png}
\pgfdeclareimage[width=\widDET]{Dhpf}{\pngDET 004_RGB__HPF_RGB.png}
\pgfdeclareimage[width=\widDET]{Dfive}{\pngDET 004_RGB_PANIGARSS_RGB.png}
\pgfdeclareimage[width=\widDET]{Dprop}{\pngDET 004_RGB_PANS5HBTY_RGB.png}

\pgfdeclareimage[width=\widDET]{Fx}{\pngFULL 004_RGB_bicubic_RGB.png}
\pgfdeclareimage[width=\widDET]{Fz}{\pngFULL 004_RGB_RGB.png}
\pgfdeclareimage[width=\widDET]{Fhpf}{\pngFULL 004_RGB__HPF_band_RGB.png}
\pgfdeclareimage[width=\widDET]{Ffive}{\pngFULL 004_RGB_PANIGARSS_RGB.png}
\pgfdeclareimage[width=\widDET]{Fprop}{\pngFULL 004_RGB_PANS5HBTY_RGB.png}

\newcommand{\ru}{\rule{0mm}{3mm}}

\title{Advances on CNN-based super-resolution of Sentinel-2 images}
\name{Massimiliano Gargiulo}
\address{DIETI, University Federico II, Via Claudio 21, 80125, Naples (I)}

\begin{document}

\maketitle
\begin{abstract}
Thanks to their temporal-spatial coverage and free access,
Sentinel-2 images are very interesting for the community.
However, a relatively coarse spatial resolution, compared to that of
state-of-the-art commercial products, 
motivates the study of super-resolution techniques to mitigate such a limitation.
Specifically, 
thirtheen bands are sensed simultaneously but at different spatial resolutions:
10, 20, and 60 meters depending on the spectral location.
Here, building upon our previous convolutional neural network (CNN) based
method \cite{Gargiulo2018a}, 
we propose an improved CNN solution to super-resolve the 20-m resolution bands 
benefiting spatial details conveyed by the accompanying 10-m spectral bands.
\end{abstract}
\section{Introduction}
\label{sec:intro}

Data fusion is a topic of interest for the remote sensing community
arising in such diverse formulations as cross-sensor feature extraction \cite{Scarpa2018, Errico2014},
multitemporal analysis \cite{Gaetano2014},
multiresolution fusion \cite{Masi2016, Gargiulo2018a, Scarpa2018a}.
Sentinel-2 images provide global acquisitions of multispectral images with a high revisit frequency, supplying data for services such as risk management (floods, forest fires, subsidence, landslide), land monitoring, food security/early warning systems, water management, soil protection and so forth \cite{Drusch2012}.
Due to technological constraints only four out of thirteen bands are provided at the highest resolution of 10 meters. 
The remaining bands are given at 20 or 60 meters. 
%One such bands is for example the SWIR, provided at 20 meters, which has proven to be very useful for several purposes \cite{xu2006modification}.
Motivated by the above consideration in \cite{Gargiulo2018a} we proposed a CNN-based data fusion technique
for the super-resolution of the 20-m short wave infrared (SWIR) band using the other higher resolution bands to recover the missing spatial details,
following the rationale behind the more familiar pansharpening problem \cite{Vivone2015, Scarpa2018a}. 

%%%% SOA
The super-resolution of an image without auxiliary bands can be addressed 
in such diverse manners ranging from a polynomial interpolation to advanced 
CNN-based modeling \cite{Dong2016,Kim2016}.
The availability of high resolution references, or guide, like in pansharpening,
allows to obtain highly accurate super-resolution, 
thanks to the fusion between the spectral information conveyed by the target image
and the spatial content gathered by the guiding band/s.
Classical multiresolution fusion methods are often categorized as component substitution (CS) or multiresolution analysis (MRA),
a survey of which can be found in \cite{Vivone2015}.
Several pansharpening methods,
based on both CS \cite{Shah2008,Tu2001}
MRA \cite{Chavez1991,Ranchin2000}, were adapted to the Sentinel-2/SWIR case and compared
in \cite{Du2016, Wang2016}.

Other very recent approaches,
both model-based \cite{Brodu2017, Lanaras2017} and data-driven CNN-based \cite{Lanaras2018}, 
aimed to super-resolve the whole Sentinel-2 dataset testify 
the relevance of the topic to the community.
%%%

In this work we propose an improved version of our previous method \cite{Gargiulo2018a} 
that counts four main integrations, (i) the use of the residual learning strategy \cite{He2016},
(ii) the batch normalization \cite{Ioffe2015}, both (i) and (ii) aimed to speed-up the learning, 
(iii) a high-pass preprocessing of the input which has already 
proven to be effective for pansharpening in \cite{Yang2017},
and, finally, (iv) the extension to all 20-m bands.

The rest of the paper is organized as follows. 
In the next section we summarize the proposed solution. 
Section 3 presents numerical and visual results. 
Finally, Section 4 draws concluding remarks.

%\vfill
%\pagebreak

\section{Proposed Sentinel-2 image super-resolution}

 \begin{figure*}[h]
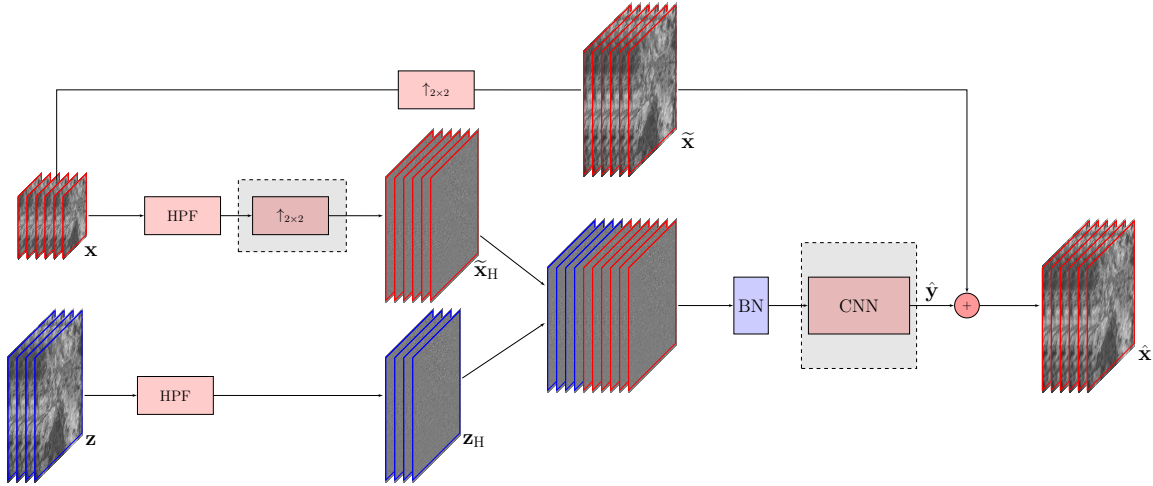

\centering
	\image{net2}
	\caption{\small Top-level workflow of the proposed super-resolution method for 20-m bands of Sentinel-2. 
	Only dashed boxes are used in \cite{Gargiulo2018a}.}
	\label{fig:flowchart}
\end{figure*}
The proposed solution is summarized in the top-level flowchart of Fig.\ref{fig:flowchart}. 
At the core is a three-layer CNN block whose internal architecture is nearly the same 
of \cite{Gargiulo2018a}, the main difference being the output shape 
which counts here six bands rather than just the single SWIR.
For the sake of brevity,
we skip the description of the internal setting of the CNN
as the Reader can refer to \cite{Gargiulo2018a} to this purpose, 
focusing on the remaining architectural aspects.

First, 
it can be noticed the additional batch normalization (BN) layer which processes the input stack $(\zH,\xH)$
that feeds the network in order to make the learning process robust with respect to the 
statistical fluctuations of the training dataset \cite{Ioffe2015}.

The BN block takes a concatenation stack as input which comprises both the high pass filtered (HPF) 10-m resolution bands of Sentinel-2, $\zH$,
and the upscaled (via bicubic interpolation) HPF version of the target 20-m resolution bands, $\xH$.
The use of HPF bands in place of the original ones has been introduced in the contex of pansharpening in \cite{Yang2017}
on the basis on the intuition that low spatial frequencies do not carry relevant information about spatial details,
therefore they can be neglected eventually facilitating the training process.

Another relevant trick generally useful when dealing with deep convolutional networks,
and actually effective for super-resolution and pansharpening as well \cite{Scarpa2018a},
is the use of residual learning \cite{He2016}, which is here implemented 
through the skip connection (on top) that directly links the input (upscaled) to the output ($\hx = \tx + \hy$).
Intuitively speaking,
since the low-frequency content of the desired output is already comprised in the low resolution
input bands, it is sufficient for the network to learn only how to predict the complementary
detail component $\hy$ to be combined with the input $\tx$ in order to get the full-resolution product.
The practical consequence of this is that the network learns much faster, 
which is very useful to finetune the network depending on the dataset.

\subsection{Training setting}

In order to train the network's parameters
a sufficiently large number of input-output examples
and the choice of a suitable cost function to minimize on them
are required to run any learning algorithm, 
like for example the Stochastic Gradient Descent
as done in \cite{Gargiulo2018a, Scarpa2018a} and here, as well,
using the $L_1$-norm as cost.
Due to the lack of ideally super-resolved samples to be used to this purpose,
it is becoming a common practice to resort to a self-supervised learning strategy
that sounds like the so-called Wald's protocol, 
a procedure commonly used to compare objectively different pansharpening algorithms
using referenced data \cite{Vivone2015}.
Given a training Sentinel-2 sample $(\bx,\bz)$, 
distinguishing between low and high resolution bands, respectively,
a training example is generated as follows.
Both bands subsets undergo a downsampling whose band-wise anti-aliasing filters
mimic the corresponding sensor characteristics (cutoff frequencies)
and the resolution downgrade factor is equal to the scale factor between $\bz$ and $\bx$,
which is $2\times2$ in our case.
Said $(\bz_\downarrow, \bx_\downarrow)$ the reduced-resolution sample so obtained, 
the original band set $\bx$ can now play the role of ideal output (reference or label)
corresponding to  $(\bz_\downarrow, \bx_\downarrow)$.
This process is repeated on the whole training image collection whose size 
must be consistent with the number of parameters to train.
The larger the network the larger must be the training dataset to avoid overfitting problems.
Additional details on training can be found in \cite{Gargiulo2018a} as this part has not been modified.

\section{Experimental Results}

\begin{table}
\centering
\def\arraystretch{0.8}
\setlength{\tabcolsep}{3pt}
\footnotesize
\begin{tabular}{l|ccccc}  
\hline
\ru \textbf{Method} & \textbf{Q-index} & \textbf{ERGAS} & \textbf{HCC} & \textbf{Epochs}   & \textbf{Sec./Ep.}\\ %& \textbf{HCI}

\ru & (1)  & (0)   & (1) & & \\

\hline
%\ru \ru bicubic (without $\bz$)  & $ 0.9825  $ & $ 5.340 $ & $0.4444 $ & - & - \\
\ru M5 \cite{Gargiulo2018a} (early stop)  & $ 0.9853 $ & $ 4.950 $ &  $ 0.5686$ & 200  & 3.577 \\
\ru M5 \cite{Gargiulo2018a}  & $ 0.9908 $ & $ 3.619 $ &  $ 0.8482$ & 1500 & 3.577 \\
\hline 

\ru Proposed (without $\bz$)    & $ 0.9905 $ & $ 3.985 $ &  $ 0.5792$ & 200  & 1.918\\
%\ru $\mathbf{HSR_{5}}$ (Ablation)    & $ 0.9926 $ & $ 2.917$ &  $ 0.7343$   & 200 & 3.566 \\
\ru Proposed (no HPF)  & $ 0.9848  $ & $ 3.112 $ &  $ 0.8482$ & 200 & 3.613\\
\ru Proposed & $\mathbf{0.9978 }$ & $\mathbf{1.937}$ & $\mathbf{0.8819} $ & 200 & 3.597\\

\hline
\end{tabular}
\caption{Ablation study}
\label{tab:ablation}
\end{table}

\begin{table}
\centering
\def\arraystretch{0.8}
\setlength{\tabcolsep}{3pt}
\footnotesize
\begin{tabular}{l|ccc}  
\hline
\ru \textbf{Method} & \textbf{Q-index} & \textbf{ERGAS} & \textbf{HCC}  \\
\ru & (1)  & (0)   & (1)  \\
\hline 
\ru bicubic (without $\bz$)  & $ 0.9825  $ & $ 5.340 $ & $0.4444 $  \\
\ru PCA  & $ 0.9596 $ & $ 8.331 $ & $0.7477 $  \\
\ru IHS  & $0.9434 $ & $ 9.544 $ & $0.7431 $  \\
\ru HPF  \cite{Chavez1991} & $ 0.9909 $ & $ 3.891 $ & $0.7091 $  \\
\ru GS2-GLP  & $ 0.9919  $ & $ 3.672 $ & $0.7414 $  \\
\ru Indusion  & $ 0.9877  $ & $ 4.494 $ & $0.6844 $  \\
\ru M5 \cite{Gargiulo2018a}  & $ 0.9908 $ & $ 3.619 $ &  $ 0.8482$\\
\ru Proposed & $\mathbf{0.9978 }$ & $\mathbf{1.937}$ & $\mathbf{0.8819} $\\
\hline

\end{tabular}
\caption{Comparative analysis.}
\label{tab:comparison}
\end{table}

In order to assess the performance of the proposed method a separate test image,
a 2700$\times$2700 scene (about Rome city center),
not used in the training phase,
has been considered.
The numerical accuracy was assessed by means of commonly used indicators like the
quality (Q-)index and ERGAS (see \cite{Vivone2015} for a detailed description of these measures),
in addition to HCC \cite{Gargiulo2018a}, a correlation coefficient computed over detail components.
Comparative methods are the bicubic interpolator (it does not take any auxiliary high-resolution guide $\bz$ as input),
just as baseline, the model M5 proposed in \cite{Gargiulo2018a} properly generalized to output all six bands $\hx$,
plus other classical pansharpening methods (which can be found in \cite{Vivone2015}) suitably adapted to the case of Sentinel-2.

In Tab.\ref{tab:ablation} it is summarized an ablation study restricted to the proposed method and \cite{Gargiulo2018a}.
A first observation is about the training cost in terms of computational time (last two columns). 
The tradeoff between training time and accuracy is highlighted for M5 that requires 
1500 epochs to reach an accuracy closer to that obtained by the proposed model trained with only 200 epochs.
Focusing on the proposal
it can be noticed, as expected, that by using additional guiding high-resolution bands $\bz$ is certainly beneficial.
Moreover, the introduction of the high pass filtering provides further improvements, notably in terms of ERGAS.

Tab.\ref{tab:comparison} gathers the numerical comparison between the proposed solution and the reference methods.
The proposal clearly outperforms all comparative solutions with respect to all quality indicators considered here.

%%% VISUAL COMPARISON

\begin{figure}
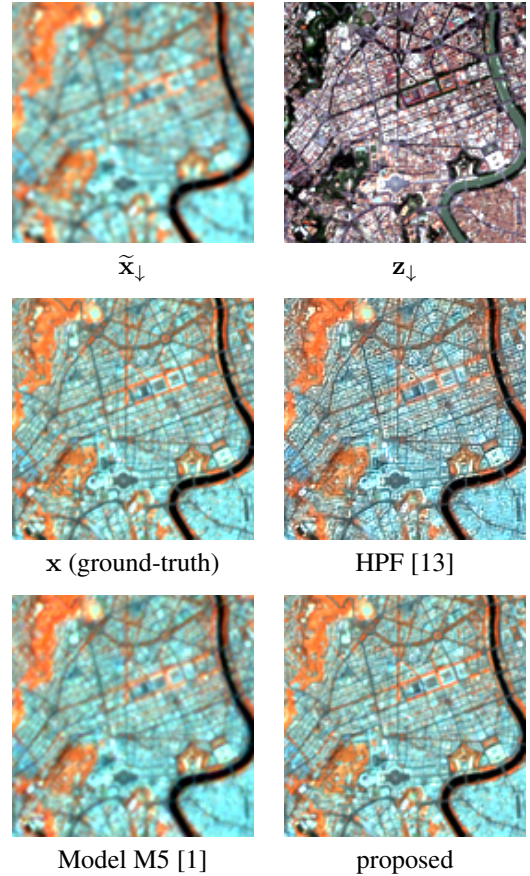

\centering
\begin{tabular}{cc}
\image{Bx} & \image{Bz} \\
$\tx_\downarrow$ & $\bz_\downarrow$ \\[2mm]
\image{Br} & \image{Bhpf} \\
$\bx$ (ground-truth) & HPF \cite{Chavez1991}\\[2mm]
\image{Bfive} & \image{Bprop} \\
Model M5 \cite{Gargiulo2018a} & proposed
\end{tabular}
\caption{Super-resolution results on a 130$\times$130 tile of the test image (Rome), in the reduced-resolution domain.}
\label{fig:BIG}
\end{figure}

\begin{figure}
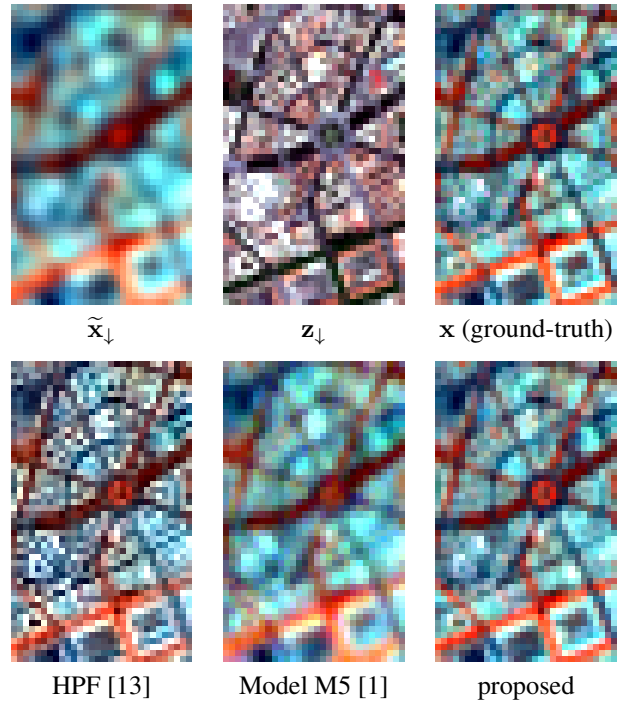

\centering
\begin{tabular}{ccc}
\image{Dx} & \image{Dz} & \image{Dr} \\
$\tx_\downarrow$ & $\bz_\downarrow$ & $\bx$ (ground-truth)\\[2mm]
\image{Dhpf} & \image{Dfive} & \image{Dprop} \\
 HPF \cite{Chavez1991} & Model M5 \cite{Gargiulo2018a} & proposed
\end{tabular}
\caption{Zoomed detail in the reduced-resolution domain.}
\label{fig:DET}
\end{figure}

To complete the analysis of the results let us look at some sample result.
For the sake of brevity we only show a limited portion of the Rome image,
selecting a meaningful subset of comparative solutions.
A 130$\times$130 cropped sample is shown in Fig.\ref{fig:BIG}.
Here the reduced resolution behaviour is analysed so that we can benefit 
from the availability of a ground-truth. 
For the purpose of visualization only three out of 6 (4) bands of $\tx_\downarrow$ ($\bz_\downarrow$) are shown 
in a false color RGB representation. 
Likewise, truth $\bx$ and compared results are shown according with the same restriction.
On the top row is the input, split in low ($\tx_\downarrow$) and high ($\bz_\downarrow$) resolution components.
In the middle row are the ideal super-resolution and HPF \cite{Chavez1991}.
On the bottom are gathered M5 \cite{Gargiulo2018a} and the proposal.
From visual inspection it can be appreciated the fidelity of the proposed solution in comparison to the reference methods.
To further highlight the different behaviours a zoomed detail is shown in Fig.\ref{fig:DET} 
whose discussion is skipped leaving the Reader to make his/her own subjective evaluation.

\begin{figure}
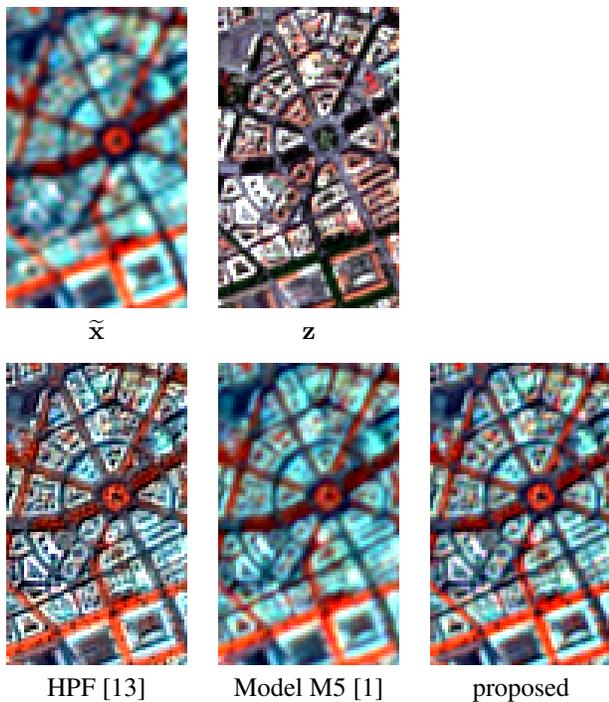

\centering
\begin{tabular}{ccc}
\image{Fx} & \image{Fz} &  \\
$\tx$ & $\bz$ & \\[2mm]
\image{Fhpf} & \image{Ffive} & \image{Fprop} \\
 HPF \cite{Chavez1991} & Model M5 \cite{Gargiulo2018a} & proposed
\end{tabular}
\caption{Zoomed detail in the full-resolution domain.}
\label{fig:FULL}
\end{figure}

Finally, in Fig.\ref{fig:FULL} we show the full-resolution domain results corresponding to the same area
selected in Fig.\ref{fig:DET}. Although we do not have a ground-truth in this case,
the different behaviours among compared solutions can be easily noticed.

\section{Conclusion}
In this work we have proposed a novel CNN-based fusion method designed to double the resolution of the 20-m bands of Sentinel-2
by taking advantage of the companion higher resolution 10-m subset of bands.
The achieved results look quite promising encouraging us to further investigate on this research line in the near future.
Of a particular interest it would be the collection of a much larger training dataset that would enable to reliably train much deeper networks.

\bibliographystyle{IEEEbib}
\newcommand*{\bibfont}{\footnotesize}
\bibfont{\bibliography{refs}}

\begin{thebibliography}{10}

\bibitem{Gargiulo2018a}
M.~Gargiulo, A.~Mazza, R.~Gaetano, G.~Ruello, and G.~Scarpa,
\newblock ``A {CNN}-based fusion method for super-resolution of {S}entinel-2
  data,''
\newblock in {\em IGARSS 2018 - 2018 IEEE International Geoscience and Remote
  Sensing Symposium}, July 2018, pp. 4713--4716.

\bibitem{Scarpa2018}
G.~Scarpa, M.~Gargiulo, A.~Mazza, and R.~Gaetano,
\newblock ``A {CNN-Based Fusion Method for Feature Extraction from Sentinel
  Data},''
\newblock {\em Remote Sensing}, vol. 10, no. 2, pp. 236, 2018.

\bibitem{Errico2014}
A.~Errico, C.~V. Angelino, L.~Cicala, D.~P. Podobinski, G.~Persechino,
  C.~Ferrara, M.~Lega, A.~Vallario, C.~Parente, G.~Masi, R.~Gaetano, G.~Scarpa,
  D.~Amitrano, G.~Ruello, L.~Verdoliva, and G.~Poggi,
\newblock ``{SAR/multispectral} image fusion for the detection of environmental
  hazards with a gis,''
\newblock in {\em Proceedings of SPIE - The International Society for Optical
  Engineering}, 2014, vol. 9245.

\bibitem{Gaetano2014}
R.~Gaetano, D.~Amitrano, G.~Masi, G.~Poggi, G.~Ruello, L.~Verdoliva, and
  G.~Scarpa,
\newblock ``Exploration of multitemporal {COSMO-skymed} data via interactive
  tree-structured {MRF} segmentation,''
\newblock {\em IEEE Journal of Selected Topics in Applied Earth Observations
  and Remote Sensing}, vol. 7, no. 7, pp. 2763--2775, 2014.

\bibitem{Masi2016}
G.~Masi, D.~Cozzolino, L.~Verdoliva, and G.~Scarpa,
\newblock ``Pansharpening by convolutional neural networks,''
\newblock {\em Remote Sensing}, vol. 8, no. 7, pp. 594, 2016.

\bibitem{Scarpa2018a}
G.~Scarpa, S.~Vitale, and D.~Cozzolino,
\newblock ``Target-adaptive {CNN}-based pansharpening,''
\newblock {\em IEEE Transactions on Geoscience and Remote Sensing}, vol. 56,
  no. 9, pp. 5443--5457, Sep. 2018.

\bibitem{Drusch2012}
{M. Drusch et al.},
\newblock ``Sentinel-2: Esa's optical high-resolution mission for gmes
  operational services,''
\newblock {\em Remote Sensing of Environment}, vol. 120, no. Supplement C, pp.
  25 -- 36, 2012,
\newblock The Sentinel Missions - New Opportunities for Science.

\bibitem{Vivone2015}
G.~Vivone, L.~Alparone, J.~Chanussot, M.~Dalla Mura, A.~Garzelli, G.~A.
  Licciardi, R.~Restaino, and L.~Wald,
\newblock ``A critical comparison among pansharpening algorithms,''
\newblock {\em IEEE Trans. Geosci. Remote Sens.}, vol. 53, no. 5, pp.
  2565--2586, May 2015.

\bibitem{Dong2016}
C.~Dong, C.C. Loy, K.~He, and X.~Tang,
\newblock ``Image super-resolution using deep convolutional networks,''
\newblock {\em IEEE Transactions on Pattern Analysis and Machine Intelligence},
  vol. 38, no. 2, pp. 295--307, Feb 2016.

\bibitem{Kim2016}
J.~K.~Lee J.~Kim and K.~M. Lee,
\newblock ``Accurate image super-resolution using very deep convolutional
  networks,''
\newblock in {\em Computer Vision and Pattern Recognition (CVPR), 2016 IEEE
  Conference on}, 2016, pp. 1646--1654.

\bibitem{Shah2008}
V.~P. Shah, N.~H. Younan, and R.~L. King,
\newblock ``An efficient pan-sharpening method via a combined adaptive pca
  approach and contourlets,''
\newblock {\em IEEE Transactions on Geoscience and Remote Sensing}, vol. 46,
  no. 5, pp. 1323--1335, May 2008.

\bibitem{Tu2001}
Te-Ming Tu, Shun-Chi Su, Hsuen-Chyun Shyu, and Ping~S. Huang,
\newblock ``A new look at ihs-like image fusion methods,''
\newblock {\em Information Fusion}, vol. 2, no. 3, pp. 177 -- 186, 2001.

\bibitem{Chavez1991}
P.S. Chavez and J.A. Anderson,
\newblock ``{Comparison of three different methods to merge multiresolution and
  multispectral data: Landsat TM and SPOT panchromatic},''
\newblock {\em Photogrammetric Engineering and Remote Sensing}, vol. 57, no. 3,
  pp. 295 -- 303, 1991.

\bibitem{Ranchin2000}
T.~Ranchin and L.~Wald,
\newblock ``{Fusion of high spatial and spectral resolution images: the ARSIS
  concept and its implementation},''
\newblock {\em Photogrammetric engineering and remote sensing}, vol. 66, no. 1,
  pp. 49--61, 2000.

\bibitem{Du2016}
Y.~Du, Y.~Zhang, F.~Ling, Q.~Wang, W.~Li, and X.~Li,
\newblock ``Water bodies’ mapping from sentinel-2 imagery with modified
  normalized difference water index at 10-m spatial resolution produced by
  sharpening the swir band,''
\newblock {\em Remote Sensing}, vol. 8, no. 4, pp. 354, 2016.

\bibitem{Wang2016}
Qunming Wang, Wenzhong Shi, Zhongbin Li, and Peter~M. Atkinson,
\newblock ``Fusion of sentinel-2 images,''
\newblock {\em Remote Sensing of Environment}, vol. 187, pp. 241 -- 252, 2016.

\bibitem{Brodu2017}
N.~Brodu,
\newblock ``Super-resolving multiresolution images with band-independent
  geometry of multispectral pixels,''
\newblock {\em IEEE Transactions on Geoscience and Remote Sensing}, vol. 55,
  no. 8, pp. 4610--4617, Aug 2017.

\bibitem{Lanaras2017}
Charis Lanaras, Jose Bioucas-Dias, Emmanuel Baltsavias, and Konrad Schindler,
\newblock ``Super-resolution of multispectral multiresolution images from a
  single sensor,''
\newblock in {\em The IEEE Conference on Computer Vision and Pattern
  Recognition (CVPR) Workshops}, July 2017.

\bibitem{Lanaras2018}
``Super-resolution of sentinel-2 images: Learning a globally applicable deep
  neural network,''
\newblock {\em ISPRS Journal of Photogrammetry and Remote Sensing}, vol. 146,
  pp. 305 -- 319, 2018.

\bibitem{He2016}
K.~He, X.~Zhang, S.~Ren, and J.~Sun,
\newblock ``Deep residual learning for image recognition,''
\newblock in {\em 2016 IEEE Conference on Computer Vision and Pattern
  Recognition (CVPR)}, June 2016, pp. 770--778.

\bibitem{Ioffe2015}
Sergey Ioffe and Christian Szegedy,
\newblock ``Batch normalization: Accelerating deep network training by reducing
  internal covariate shift,''
\newblock pp. 448--456, 2015.

\bibitem{Yang2017}
J.~Yang, X.~Fu, Y.~Hu, Y.~Huang, X.~Ding, and J.~Paisley,
\newblock ``Pannet: A deep network architecture for pan-sharpening,''
\newblock in {\em ICCV}, Oct. 2017.

\end{thebibliography}

\end{document}